\documentstyle[lingmacros,
               mylingmacros,
               fullname]{article}
\newdimen\lexentrydimen
\newdimen\lexcategorydimen
\newdimen\mothernodedimen
\def \lexentry #1#2#3{\setbox1\hbox{\it #1}\setbox2\hbox{#2}%
                      \ifdim \wd1 >3em \lexentrydimen=\wd1 %
                      \else \lexentrydimen=3em \fi%
                      \ifdim \wd2 >1em \lexcategorydimen=\wd2 %
                      \else \lexcategorydimen=1em \fi%
                      \noindent%
                      \makebox[\lexentrydimen][l]{\box1}\hspace*{1em}\nobreak%
                      \makebox[\lexcategorydimen][l]{\box2}\hspace*{1em}\nobreak%
                      \feqs{#3}}

\def \feqs #1{\begin{tabular}[t]{@{\strut}l@{\strut}}#1\end{tabular}}

\def \phraserule #1#2{\setbox1\hbox{#1}%
                      \ifdim \wd1 >2em \mothernodedimen=\wd1 %
                      \else \mothernodedimen=2em \fi%
   \leavevmode\hbox{\makebox[\mothernodedimen][l]{\box1}{$\longrightarrow$}\hspace{.6em}#2}}

\def \rulenode #1{\begin{tabular}[t]{c}#1\end{tabular}}

\def \? {\leavevmode \llap {?}}
\def \up {\hbox {$\uparrow$\kern .2em}}
\def \down {\hbox {$\downarrow$\kern 0em}}

\def \obj {\hbox {\sc obj}}   \def \OBJ {\obj}
   
\def \obl {\hbox {\sc obl}}

\def \subj {\hbox {\sc subj}}  \def \SUBJ {\subj}

\def \pred {\hbox {\sc pred}}  \def \PRED {\pred}

\def \var {\hbox{\sc var}}

\def \restr {\hbox {\sc restr}}  
  
\def \spec {\hbox {\sc spec}}

\def \spec {\hbox {\sc spec}}

\def \tense {\hbox {\sc tense}}

\def \past {\hbox {\sc past}}

\input tree-dvips

\makeatletter
\def\section{\@startsection {section}{1}{\z@}{-3.5ex plus -1ex minus
 -.2ex}{2.3ex plus .2ex}{\normalsize\bf}}
\def\subsection{\@startsection{subsection}{2}{\z@}{-3.25ex plus -1ex minus
 -.2ex}{1.5ex plus .2ex}{\normalsize\bf}}
\def\subsubsection{\@startsection{subsubsection}{3}{\z@}{-3.25ex plus
-1ex minus -.2ex}{1.5ex plus .2ex}{\normalsize\bf}}
\def\paragraph{\@startsection
 {paragraph}{4}{\z@}{3.25ex plus 1ex minus .2ex}{-1em}{\normalsize\bf}}

\newcommand{\linimp}{\;\mbox{$-\hspace*{-.4ex}\circ$}\;}
\newcommand{\means}{\makebox[1.2em]{$\leadsto$}}
\newcommand{\meansub}[1]{\makebox[1.5em]{$\leadsto_{#1}$}}
\newcommand{\IT}[1]{\mbox{\it #1\/}}
\newcommand{\BF}[1]{\mbox{\bf #1}}
\newcommand{\pex}[1]{(\ref{#1})}
\newcommand{\attr}[2]{\mbox{$(#1\;#2)$}}
\newcommand{\Up}{\mbox{$\uparrow$}}
\newcommand{\Down}{\mbox{$\downarrow$}}
\newcommand{\Ups}{\Up_\sigma}
\newcommand{\All}[1]{\forall #1.\;}

\def\fd#1{\setlength{\baselineskip}{0pt}
  \small
     \hbox{#1}}
\def\fdx#1{\setlength{\baselineskip}{0pt}\vcenter{#1}}
\def\fdand#1{\(\left[\fdx{#1}\right]\)}

\def\feat#1#2{\vskip
.4ex\hbox{\hspace{.2em}#1\hspace{1em}#2\hspace{.2em}}\vskip .8ex}

\def\var{\mbox{\sc var}}
\def\restr{\mbox{\sc restr}}
\def\ant{\mbox{\sc ant}}
\def\alt{{\tt\char`\|}}
\makeatother

\title{A Deductive Account of Quantification in LFG\thanks{Xerox Palo
Alto Research Center Technical Report ISTL-NLTT-1993-06-01. To appear
in {\it Quantifiers, Deduction, and Context}, ed.\ Makoto Kanazawa,
Christopher J.~Pi\~{n}\'{o}n, and Henriette de~Swart.  Stanford,
California: Center for the Study of Language and Information, 1994.}}
\author{Mary Dalrymple\thanks{Xerox PARC, Palo Alto, California.}\\
John Lamping\footnotemark[2]\\
Fernando Pereira\thanks{AT\&T Bell Laboratories, Murray Hill, New Jersey.}\\
Vijay Saraswat\footnotemark[2]}

\begin{document}
\maketitle

The relationship between Lexical-Functional Grammar (LFG) {\em
functional structures} (f-structures) for sentences and their semantic
interpretations can be expressed directly in a fragment of linear
logic in a way that explains correctly the constrained interactions
between quantifier scope ambiguity and bound anaphora.

The use of a deductive framework to account for the compositional
properties of quantifying expressions in natural language obviates the
need for additional mechanisms, such as Cooper storage, to represent
the different scopes that a quantifier might take.  Instead, the
semantic contribution of a quantifier is recorded as an ordinary
logical formula, one whose use in a proof will establish the scope of
the quantifier.  The properties of linear logic ensure that each
quantifier is scoped exactly once.

Our analysis of quantifier scope can be seen as a recasting of
Pereira's analysis \cite{Pereira:HOD}, which was expressed in
higher-order intuitionistic logic.  But our use of LFG and linear
logic provides a much more direct and computationally more
flexible interpretation mechanism for at least the same range of
phenomena.  We have developed a preliminary Prolog implementation
of the linear deductions described in this work.

\section{Introduction}

This paper describes part of our ongoing investigation on the use of
formal deduction in linear logic to explicate the relationship between
syntactic analyses in Lexical-Functional Grammar (LFG) and semantic
interpretations. The use of formal deduction in semantic
interpretation was implicit in deductive systems for categorial syntax
\cite{Lambek:SentStruct}, and has been made explicit through
applications of the Curry-Howard parallelism between proofs and terms
in more recent work on categorial semantics
\cite{vanBenthem:lambek,VanBenthem:LgInAction}, labeled deductive
systems \cite{Moortgat:labelled} and flexible categorial systems
\cite{Hendriks:flexibility}. Accounts of the syntax-semantics
interface in the categorial tradition require that syntactic and
semantic analyses be formalized in parallel algebraic structures of
similar signatures, based on generalized application and abstraction
(or residuation) operators, and structure-preserving relations between
them.  Those accounts therefore force the adoption of categorial
syntactic analyses, with their strong dependence on phrase structure
and linear order.

In contrast, our approach uses linear logic \cite{Girard:Linear} to
represent the connection between two dissimilar levels of
representation, LFG f-structures and their semantic interpretations.
F-structures provide a crosslinguistically uniform representation of
syntactic information relevant to semantic interpretation that
abstracts away from the details of phrase structure and linear order
in particular languages. This generality is in part achieved by using
grammatical functions rather than functor-argument relations to
represent syntactic predicate-argument relationships.  As
\namecite{Halvorsen:Sitsem} notes, however, the flatter,
unordered, grammatical function structure of LFG does not fit well
with traditional semantic compositionality, based on functional
abstraction and application, which mandates a rigid order of semantic
composition.  We are thus forced to use a more relaxed form of
compositionality, in which, as in more traditional ones, the semantics
of each lexical entry in a sentence is used exactly once in
interpretation, but without imposing a rigid order of composition. It
turns out that linear logic offers exactly what is required for a
calculus of semantic composition for LFG, in that it can represent
directly the constraints on the creation and use of semantic units in
sentence interpretation without forcing a particular hierarchical
order of composition except as required by the properties of
particular lexical entries.

We have shown previously that the linear-logic formalization of the
syntax-semantics interface for LFG provides simple and general
analyses of modification, functional completeness and coherence, and
complex predicate formation~\cite{DLS:EACL,DHLS:ROCLING}.  In the
present paper, the analysis is extended to the interpretation of
quantified noun phrases.  After an overview of the approach, we
present our analysis of the compositional properties of quantifiers,
and we conclude by showing that the analysis correctly accounts for
scope ambiguity and its interactions with bound anaphora.

\section{LFG and Linear Logic}

\paragraph{Syntactic Framework}

LFG assumes two syntactic levels of representation: constituent
structure ({\it c-structure}) represents phrasal dominance and
precedence relations, while functional structure ({\it
f-structure}) represents syntactic predicate-argument structure.
For example, the f-structure for sentence \pex{ex:bah} is given
in \pex{ex:bahfs}:

\enumsentence{\label{ex:bah} Bill appointed Hillary.}

\enumsentence{\label{ex:bahfs}\evnup{\fd{
\fdand{\feat{\pred}{`appoint'}
       \feat{\subj}{\fdand{\feat{\pred}{`Bill'}}}
       \feat{\obj}{\fdand{\feat{\pred}{`Hillary'}}}}}}}

\noindent As illustrated, a functional structure consists of a collection
of attributes, such as \pred, \subj, and \obj, whose values can, in
turn, be other functional structures.  The following annotated
phrase-structure rules can generate the f-structure in \pex{ex:bahfs}:
\enumsentence{
\phraserule{S}{
\rulenode{NP\\ \attr{\Up}{\subj} = \Down}
\rulenode{VP\\ \up = \Down}}\\
\phraserule{VP}{
\rulenode{V\\ \up = \Down}
\rulenode{NP\\ \attr{\Up}{\obj} = \Down}}}

\noindent These two phrase structure rules do not encode semantic
information; they specify only how grammatical functions such as
\subj\ are expressed in English.  The f-structure metavariables
\Up\ and \Down\ refer, respectively, to the f-structure of the mother
of the current node and to the f-structure of the current node
\cite{KaplanBresnan:LFG}.  The annotations on the S rule
indicate, then, that the f-structure for the S has a \subj\ attribute
whose value is the f-structure for the NP daughter, and that the
f-structure for the S is the same as the one for the VP daughter.  The
relation between the nodes of the c-structure and the f-structure for
the sentence \pex{ex:bah} is expressed by means of arrows in (\ex{1}):

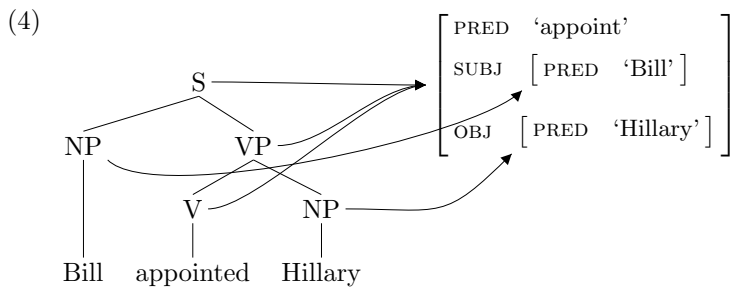
\enumsentence{\evnup{
\modsmalltree{3}{\mc{3}{\node{l}{S}\hspace*{1em}}\\
            \node{m}{NP}&\mc{2}{\hspace*{1em}\node{n}{VP}}\\
                        &\node{p}{V}&\node{t}{NP}\\
            \node{q}{Bill}&\node{r}{appointed}&\node{s}{Hillary}}%
\hspace*{2em}%
\fd{
\node{a}{\fdand{\feat{\PRED}{`appoint'}
       \feat{\SUBJ}{\node{b}{\fdand{\feat{\PRED}{`Bill'}}}}
       \feat{\ }{\ }
       \feat{\OBJ}{\node{c}{\fdand{\feat{\PRED}{`Hillary'}}}}}}}
\anodecurve[r]{p}[l]{a}{2em}
\anodecurve[r]{n}[l]{a}{2em}
\anodecurve[r]{l}[l]{a}{2em}
\anodecurve[br]{m}[bl]{b}{2.7em}
\anodecurve[r]{t}[bl]{c}{4em}
\nodeconnect{l}{m}
\nodeconnect{l}{n}
\nodeconnect{m}{q}
\nodeconnect{n}{p}
\nodeconnect{n}{t}
\nodeconnect{t}{s}
\nodeconnect{p}{r}}}

\paragraph{Lexically-Specified Semantics}

Unlike phrase structure rules, lexical entries specify semantic
as well as syntactic information.  Here are the lexical entries
for the words in the sentence:

\enumsentence{
\lexentry{Bill}{NP}{
\attr{\Up}{\pred} = `Bill'\\ $\Ups \means \IT{Bill}$}

\lexentry{appointed}{V}{
\attr{\Up}{\pred}= `appoint'\\
\hspace*{-6em}$\All{ X, Y}\attr{\Up}{\subj}_\sigma\means X \otimes
\attr{\Up}{\obj}_\sigma\means Y \linimp
                \Ups \means \IT{appoint}\/(X, Y)$}

\lexentry{Hillary}{NP}{
\attr{\Up}{\pred} = `Hillary'\\
$\Ups \means \IT{Hillary}$}\label{ex:bah-lex}}

\noindent Just like phrase structure rules, lexical entries are
instantiated for a particular utterance.  The metavariable $\Up$ in a
lexical entry represents the f-structure of the c-structure mother of
(an instance of) the entry in a c-structure.  The syntactic information
given in lexical entries consists of equality statements about the
f-structure, while the semantic information consists of assertions
about how the meaning of the f-structure participates in various
semantic relations.

The semantic information in a lexical entry, which we will call the
{\em semantic contribution} of the entry, is a linear-logic formula
that constrains the association between {\em semantic structures}\/
projected from the f-structures mentioned in the lexical entry
\cite{Kaplan:3Sed,HalvorsenKaplan:Projections} and their semantic
interpretations.  The semantic projection function $\sigma$ maps an
f-structure to a semantic structure encoding information about its
meaning, in the same way as the functional projection function $\phi$
maps c-structure nodes to the associated f-structures.  The
association between $f_\sigma$ and a meaning $P$ is represented by the
atomic formula $f_\sigma \means P$, where $\means$ is an otherwise
uninterpreted binary predicate symbol.  (In fact, we use not one but a
family of relations~$\meansub{\tau}$ indexed by the semantic type of
the intended second argument, although for simplicity we will omit the
type subscript whenever it is determinable from context.)  We will
often informally say that $P$ is $f$'s meaning without referring to
the role of the semantic structure $f_\sigma$ in $f_\sigma \means
P$. We will see, however, that f-structures and their semantic
projections must be distinguished, because in general semantic
projections carry more information than just the association to the
meaning for the corresponding f-structure.

We can now explain the semantic contributions in \pex{ex:bah-lex}.
If a particular occurrence of `Bill' in a sentence is
associated with f-structure $f$, the syntactic constraint in the
lexical entry {\em Bill} will be instantiated as
$\attr{f}{\pred} = \mbox{`Bill'}$ and the semantic constraint will be
instantiated as $f_\sigma
\means \IT{Bill}$, representing the association between $f_\sigma$ and the
constant $\IT{Bill}$ representing its meaning.

The semantic contribution of the {\em appointed} entry is more
complex, as
it relates the meanings of the subject and object of a clause
to the clause's meaning. Specifically, if $f$ is the
f-structure for a clause with predicate ($\pred$) `appoint', the
semantic contribution asserts that if $f$'s subject $\attr{f}{\subj}$ has
meaning $X$ and (linear conjunction $\otimes$) $f$'s object
$\attr{f}{\obj}$ has meaning $Y$, then (linear implication
$\linimp$) $f$ has meaning $\IT{appoint}(X,Y)$.\footnote{In fact, we believe
that the correct
treatment of the relation between a verb and its arguments requires
the use of {\it mapping principles}\/ specifying the relation between
the array of semantic arguments required by a verb and their possible
syntactic realizations
\cite{BresnanKanerva:Locative,Alsina:PhD,Butt:PhD}.  A verb like {\it
appoint}\/, for example, might specify that one of its arguments is an
agent and the other is a theme.  Mapping principles would then specify
that agents can be realized as subjects and themes as objects.

Here we make the simplifying assumption that the arguments of verbs have
already been linked to syntactic functions and that this linking is
represented in the lexicon, since for the examples we will discuss this
assumption is innocuous.  However, in the case of {\it complex predicates}
this assumption produces incorrect results, as shown by
\namecite{Butt:PhD}.  Mapping principles are very naturally incorporated
into the framework discussed here; see
\namecite{DLS:EACL} and \namecite{DHLS:ROCLING}
for discussion and illustration.}

\paragraph{Logical Representation of Semantic Compositionality}
In the semantic contribution for {\em appointed} in \pex{ex:bah-lex},
the linear-logic connectives of multiplicative conjunction $\otimes$
and linear implication $\linimp$ are used to specify how the meaning
of a clause headed by the verb is composed from the meanings of the
arguments of the verb. For the moment, we can think of the linear
connectives as playing the same role as the analogous classical
connectives conjunction and implication, but we will soon see that the
specific properties of the linear connectives are essential to
guarantee that lexical entries bring into the interpretation process
all and only the information provided by the corresponding words.  The
semantic contribution of {\em appointed} asserts that if the subject
of a clause with main verb {\em appointed} means $X$ and its object
means $Y$, then the whole clause means $\IT{appoint}(X,Y)$.  The
semantic contribution can thus be thought of as a linear definite
clause, with the variables $X$ and $Y$ playing the same role as Prolog
variables.

It is worth noting that the form of the semantic contribution of
\IT{appointed} parallels the type $e\times e \rightarrow t$ which, in
its curried form \mbox{$e\rightarrow{e}\rightarrow{t}$}, is the standard type
for a transitive verb in a compositional semantics setting
\cite{LTFGamut:vol2}. In general, the propositional structure of the
semantic contributions of lexical entries will parallel the types assigned
to the meanings of the same words in compositional analyses.

Given the semantic contributions in \pex{ex:bah-lex}, we can derive
deductively the meaning for example (\ref{ex:bah}). Let the constants
$f$, $g$ and $h$ name the following f-structures:

\enumsentence{\label{ex:bahafs}\evnup{\fd{
$f$:\fdand{\feat{\pred}{`appoint'}
       \feat{\subj}{$g$:\fdand{\feat{\pred}{`Bill'}}}
       \feat{\obj}{$h$:\fdand{\feat{\pred}{`Hillary'}}}}}}}

\noindent Instantiating the lexical entries for {\em Bill}, {\em
Hillary}, and {\em appointed} appropriately, we obtain the following
semantic contributions, abbreviated as \BF{bill}, \BF{hillary}, and
\BF{appointed}:
\[
\begin{array}{@{\strut}ll@{\strut}}
\BF{bill}\colon& g_{\sigma} \means \IT{Bill}\\
\BF{hillary}\colon& h_{\sigma} \means \IT{Hillary}\\
\BF{appointed}\colon& \All{ X, Y} g_{\sigma}\means X \otimes h_{\sigma}\means Y
\linimp
                f_{\sigma}\means \IT{appoint}(X, Y)
\end{array}
\]

\noindent These formulas show how the generic semantic contributions in
the lexical entries are instantiated to reflect their participation in
this particular f-structure.  For example, since the entry {\em Bill}
is used for f-structure $g$, the semantic contribution for {\em Bill}
provides a meaning for $g_{\sigma}$. More interestingly, the verb {\em
appointed}\/ requires two pieces of information, the meanings of its
subject and object, in no particular order, to produce a meaning for
the clause.  As instantiated, the f-structures corresponding to the
subject and object of the verb are $g$ and $h$, respectively, and $f$
is the f-structure for the entire clause.  Thus, the instantiated
entry for {\em appointed} shows how to combine a meaning for
$g_{\sigma}$ (its subject) and $h_{\sigma}$ (its object) to generate a
meaning for $f_{\sigma}$ (the entire clause).

In the following, assume that the formula \BF{bill-appointed} is
defined thus:
\[\begin{array}[t]{ll}
\BF{bill-appointed}\colon&
\All{ Y}h_{\sigma}\means Y \linimp f_{\sigma} \means \IT{appoint}(\IT{Bill}, Y)
\end{array}
\]
\noindent Then the following derivation is possible in linear logic ($\vdash$
stands
for the linear-logic entailment relation):
\enumsentence{\label{ex:bahderiv}
$
\begin{array}[t]{l@{\hspace*{2em}}l}
&\BF{bill} \otimes \BF{hillary} \otimes \BF{appointed} \quad (Premises.)
\\[.5ex]
\vdash &  \BF{bill-appointed} \otimes \BF{hillary}\\[0.5ex]
\vdash & f_{\sigma} \means appoint(Bill, Hillary)
\end{array}
$}
\noindent At each step, universal instantiation and modus ponens are used.

In summary, each word in a sentence contributes a linear-logic formula
relating the semantic projections of specific f-structures in the LFG
analysis to representations of their meanings.  From those formulas,
the interpretation process attempts to deduce an atomic formula
relating the semantic projection of the whole sentence to a
representation of the sentence's meaning. Alternative derivations may
yield different such conclusions, corresponding to semantic
interpretation ambiguities.

\paragraph{Meaning and glue}

Our approach shares the order-independence of representations of
semantic information by attribute-value matrices
\cite{PollardSag:HPSG1,FenstadEtAl:SitLgLogic,PollardSag:HPSG2}, while
still allowing a well-defined treatment of variable binding and scope.
We do this by distinguishing (1) a {\em language of meanings}\/ and (2) a
language for assembling meanings or {\em glue language}.

The language of meanings could be that of any appropriate logic, for
instance Montague's intensional logic \cite{Montague:PTQ}.  The glue
language, described below, is a fragment of linear logic.  The
semantic contribution of each lexical entry is represented by a
linear-logic formula that can be understood as instructions in the
glue language for combining the meanings of the lexical entry's
syntactic arguments into the meaning of the f-structure headed by the
entry.  Glue formulas may also be contributed by some syntactic
constructions, when properties of a construction as a whole and not
just of its lexical elements are responsible for the interpretation of
the construction; these cases include the semantics of relative
clauses.  We will not discuss construction-specific interpretation
rules in this paper.

Appendix \ref{syn-app} gives further details on the syntax of the
meaning and glue languages used in this paper.

\paragraph{Linear logic} As we have just outlined, we
use deduction in linear logic to assign meanings to sentences,
starting from information about their functional structure and about
the semantics of the words they contain.  An approach based on linear
logic, which crucially allows premises to commute, appears to be more
compatible with the shallow and relatively free-form functional
structure than are compositional approaches, which rely on deeply
nested binary-branching immediate dominance relationships.  As noted
above, the use of linear logic as the system for assembling meanings
permits a uniform treatment of a range of natural language phenomena
described by \namecite{DLS:EACL}, including modification, completeness
and coherence,\footnote{``An f-structure is {\em locally complete}\/ if
and only if it contains all the governable grammatical functions that
its predicate governs.  An f-structure is {\em complete}\/ if and only
if all its subsidiary f-structures are locally complete. An
f-structure is {\em locally coherent}\/ if and only if all the
governable grammatical functions that it contains are governed by a
local predicate.  An f-structure is {\em coherent}\/ if and only if
all its subsidiary f-structures are locally coherent.''
\cite[pages~211--212]{KaplanBresnan:LFG}.} and complex predicate
formation.

An important motivation for using linear logic is that it allows us to
to capture directly the intuition that lexical items and phrases each
contribute exactly once to the meaning of a sentence.  As noted by
\namecite[page~172]{KleinSag:Type}:
\begin{quote}
Translation rules in Montague semantics have the property that the
translation of each component of a complex expression occurs exactly
once in the translation of the whole.  \ldots That is to say, we do
not want the set S [of semantic representations of a phrase] to
contain {\em all} meaningful expressions of IL which can be built up
from the elements of S, but only those which use each element exactly
once.
\end{quote}

\noindent In our terms, the semantic contributions of the constituents
of a sentence are not context-independent assertions that may be used
or not in the derivation of the meaning of the sentence depending on
the course of the derivation. Instead, the semantic contributions are
{\em occurrences} of information which are generated and used exactly
once.  For example, the formula $g_{\sigma}\means
\IT{Bill}$ can be thought of as providing one occurrence of the meaning
$\IT{Bill}$ associated to the semantic projection $g_{\sigma}$.  That
meaning must be consumed exactly once (for example, by \BF{appointed} in
\pex{ex:bahderiv}) in the derivation of a meaning of the entire utterance.

It is this ``resource-sensitivity'' of natural language semantics---an
expression is used exactly once in a semantic derivation---that linear
logic can model. The basic insight underlying linear logic is that
logical formulas are {\em resources} that are produced and consumed in
the deduction process.  This gives rise to a resource-sensitive notion
of implication, the {\em linear implication} $\linimp$: the formula $A
\linimp B$ can be thought of as an action that can {\em consume} (one
copy of) $A$ to produce (one copy of) $B$. Thus, the formula $A
\otimes (A \linimp B)$ linearlyentails $B$.  It does not entail $A
\otimes B$ (because the deduction consumes $A$), and it does not entail
$(A \linimp B) \otimes B$ (because the linear implication is also
consumed in doing the deduction).  This resource-sensitivity not only
disallows arbitrary duplication of formulas, but also disallows
arbitrary deletion of formulas. Thus the linear multiplicative
conjunction $\otimes$ is sensitive to the multiplicity of formulas: $A
\otimes A$ is not equivalent to $A$ (the former has two copies of the
formula $A$).  For example, the formula $A \otimes A \otimes (A
\linimp B)$  linearly entails $A \otimes B$ (there is still one $A$
left over) but does not entail $B$ (there must still be one
$A$ present).  In this way, linear logic checks that a formula is used
once and only once in a deduction, enforcing the requirement that
each component of an utterance contributes exactly once to the
assembly of the utterance's meaning.

To handle quantification, our glue language needs to be only a
fragment of higher-order linear logic, the {\em tensor fragment}, that
is closed under conjunction, universal quantification, and implication
(with at most one level of nesting of implication in antecedents).  In
fact, all but the determiner lexical entries are in the first-order
subset of this fragment.  This fragment arises from transferring to
linear logic the ideas underlying the concurrent constraint
programming scheme of \namecite{Saraswat:PhD}. An explicit formulation
for the higher-order version of the linear concurrent constraint
programming scheme is given in
\namecite{SaraswatLincoln:HLCC}.  A nice tutorial introduction
to linear logic itself may be found in
\namecite{Scedrov:Linear}; see also
\namecite{Saraswat:IntroLCC}.

\paragraph{Relationship with Categorial Syntax and Semantics}

As suggested above, there are interesting connections between our
approach and various systems of categorial syntax and semantics. The
Lambek calculus
\cite{Lambek:SentStruct}, introduced as a logic of syntactic
combination, turns out to be a fragment of noncommutative
multiplicative linear logic.  If permutation is added to Lambek's
system, its left- and right-implication connectives ($\setminus$ and
$/$) collapse into a single implication connective with behavior
identical to $\linimp$. This undirected version of the Lambek calculus
was developed by
van Benthem \shortcite{vanBenthem:lambek,VanBenthem:LgInAction} to account
for the semantic combination possibilities of phrase meanings.

Those systems and related ones
\cite{Moortgat:categorial,Hepple:PhD,Morrill:intensional} were
developed as calculi of syntactic/semantic types, with propositional
formulas representing syntactic categories or semantic types. Given
the types for the lexical items in a sentence as assumptions, the
sentence is syntactically well-formed in the Lambek calculus if the
type of the sentence can be derived from the assumptions arranged as
an ordered list. Furthermore, the Curry-Howard isomorphism between
proofs and terms \cite{Howard:construction} allows the extraction of a
term representing the meaning of the sentence from the proof that the
sentence is well-formed \cite{vanBenthem:cat-lambda}. However, the
Lambek calculus and its variants carry with them a particular view of
syntactic structure that is not obviously compatible with the flatter
f-structures proposed by LFG.

On the other hand, categorial semantics in the undirected Lambek
calculus and other related commutative calculi provides an analysis of
the possibilities of meaning combination independently of the
syntactic realizations of those meanings, but does not provide a
mechanism for relating semantic combination possibilities to
the corresponding syntactic combination possibilities.

In more recent work, multidimensional and labeled deductive systems
\cite{Moortgat:labelled,Morrill:type-logical} have been proposed as
refinements of the Lambek systems that are able to represent
synchronized derivations involving multiple levels of representation,
for instance a level of head-dependent representations and a level of
syntactic functor-argument representations. However, these systems do
not yet seem able to represent the connection between a flat syntactic
representation in terms of grammatical functions and a
function-argument semantic representation. As far as we can see, the
problem in those systems is that at the type level it is not possible
to express the link between particular syntactic structures
(f-structures in our case) and particular contributions to
meaning. The extraction of meanings from derivations following the
Curry-Howard isomorphism that is standard in categorial systems
demands that the order of syntactic combination coincide with the
order of semantic combination so that functor-argument relations at
the syntactic and semantic level are properly aligned.

Thus, while the ``propositional skeleton'' of an analysis in our
system can be seen as a close relative of the corresponding categorial
semantics derivation in the undirected Lambek calculus, the
first-order part of our analysis (notably the $f$, $g$, and $h$ in the
example above) explicitly carries the connection between f-structures
and their contributions to meaning. In this way, we can take advantage
of the principled description of potential meaning combinations of
categorial semantics without losing track of the constraints imposed
by syntax on the possible combinations of those meanings.

\section{Quantification}

Our treatment of quantification, and in particular of quantifier scope
ambiguity and of the interactions between scope and bound anaphora,
follows the approach of Pereira
\shortcite{Pereira:SemComp,Pereira:HOD}.  It turns out, however, that
the linear-logic formulation is simpler and easier to justify than
the earlier analysis, which used an intuitionistic type assignment
logic.

The basic idea for the analysis can be seen as a logical
counterpart at the glue level of the standard type
assignment for generalized quantifiers
\cite{Barwise+Cooper:generalized}. The generalized quantifier
meaning of a natural language determiner has the following type, a
function from two properties, the quantifier's restriction and scope,
to a proposition:
\enumsentence{$(e\rightarrow
t)\rightarrow (e\rightarrow t) \rightarrow t$\label{eq:gen-quant-type}}
\noindent At the semantic glue level, we can understand
that type as follows. For any determiner, if for arbitrary $x$ we can
construct a meaning $R x$ for the quantifier's restriction, and again
for arbitrary $x$ we can construct a meaning $S x$ for the
quantifier's scope, where $R$ and $S$ are properties (functions from
entities to propositions), then we can construct the meaning $Q R S$
for the whole sentence containing the determiner, where $Q$ is the
meaning of the determiner.  In the following we
will notate $Q R S$ meaning more perspicuously as $Q(z,Rz,Sz)$.

Assume that we have determined the following semantic structures:
$\IT{restr}$ for the restriction (a common noun phrase),
$\IT{restr-arg}$ for its implicit argument, $\IT{scope}$ for the scope
of quantification, and $\IT{scope-arg}$ for the grammatical function
filled by the quantified noun phrase.  Then the foregoing analysis can
be represented in linear logic by the following schematic formula:
\enumsentence{$\begin{array}[t]{r@{\,}l}
\All{ R, S} & (\All{ x}\IT{restr-arg} \means x \linimp \IT{restr} \means R x)\\
    \otimes & (\All{ x}\IT{scope-arg} \means x \linimp \IT{scope} \means S x)\\
    \linimp & \IT{scope} \means Q(z, Rz, Sz)
\end{array}$\label{eq:gen-quant-lin}}
Given the equivalence between $A\otimes B\linimp C$ and $A\linimp
(B\linimp C)$, the propositional part of \pex{eq:gen-quant-lin}
parallels the generalized quantifier type \pex{eq:gen-quant-type}.

In addition to providing a semantic type assignment for determiners,
\pex{eq:gen-quant-lin} uses glue language quantification to express how
the meanings of the restriction and scope of quantification are
determined and combined into the meaning of the quantified clause.
The condition \mbox{$\All{ x}\IT{restr-arg} \means x \linimp
\IT{restr} \means R x$} specifies that, if for arbitrary $x$ $\IT{restr-arg}$
has meaning $x$,\footnote{We use lower-case
letters for {\em essentially universal} variables, that is, variables
that stand for new local constants in a proof. We use capital letters for
{\em
essentially existential} variables, that is,
Prolog-like variables that become instantiated to
particular terms in a proof. In other words, essentially existential
variables stand for specific but as yet unspecified terms, while
essentially universal variables stand for arbitrary constants, that
is, constants that could be replaced by {\em any} term while still
maintaining the validity of the derivation. In the linear-logic
fragment we use here, essentially existential variables arise from
universal quantification with outermost scope, while essentially
universal variables arise from universal quantification whose scope is
a conjunct in the antecedent of an outermost implication.}
then $\IT{restr}$ has meaning $R x$,
that is, it gives the dependency of the meaning of a common noun
phrase on its implicit argument. Property $R$ is the representation of
that dependency as a function. Similarly, the subformula \mbox{$\All{
x}\IT{scope-arg} \means x \linimp \IT{scope} \means S x$} specifies
the dependency of the meaning $S x$ of a semantic structure
$\IT{scope}$ on the meaning $x$ of one of its arguments
$\IT{scope-arg}$. If both dependencies hold, then $R$ and $S$ are an
appropriate restriction and scope for the determiner meaning $Q$.

Computationally, the nested universal quantifiers substitute unique
new constants (eigenvariables) for the quantified variable $x$, and
the nested implications try to prove their consequent with the
antecedent added to the current set of assumptions. Higher-order
unification (in this case, a fairly restricted version thereof
\cite{Miller:LLambda}) is used to solve for the values of $R$ and $S$
that satisfy the nested implication, which cannot contain occurrences
of the eigenvariables.

To complete our specification of the semantic contribution of a
determiner, we need to see how it relates to the f-structure it
contributes to.  The f-structure for a quantified noun phrase has the
general form
\enumsentence{\evnup{
\fd{$f$:\fdand{\feat{\spec}{$q$}
                 \feat{\pred}{$g$:$\cdots$}}}}}
where \spec\ is the determiner and \pred\ is the noun.

Since the meaning of a noun is a property (type $e\rightarrow t$), its
semantic contribution has the form of an implication, just like a
verb.\footnote{For a discussion of relational nouns, whose meanings
are relations rather than properties, see Section
\ref{sec:constraints}.}  However, while for a verb the arguments and
result of the verb meaning can be associated to projections of
appropriate f-structures in the syntactic analysis of the verb's
clause, there are no appropriate f-structures in the analysis of a
noun phrase that can be associated with the argument and result of the
noun's meaning.  Instead, we take the semantic projection $f_\sigma$
of the noun phrase to be structured with two attributes
$\attr{f_\sigma}{\var}$ and $\attr{f_\sigma}{\restr}$, performing a
comparable role to the attributes {\sc cm}\ and {\sc pm}\ in the
semantic structure in Halvorsen's \shortcite{Halvorsen:LI} treatment
of quantifiers.  Using those attributes, the semantic contribution of
a noun can be expressed in the form $$\All{ x}
\attr{\Ups}{\var}
\means x \linimp \attr{\Ups}{\restr} \means P x$$ where $P$ is the
meaning of the noun.

We can now describe how the semantic structures $\IT{restr-arg}$,
$\IT{restr}$, $\IT{scope-arg}$ and $\IT{scope}$ in
\pex{eq:gen-quant-lin} relate to the f-structure.
The contribution of determiner $q$ is expressed in terms of $f$'s
semantic projection $f_\sigma$.  To connect the restriction of the
determiner with the noun, we take
$\IT{restr-arg}=\attr{f_\sigma}{\var}$ and
$\IT{restr}=\attr{f_\sigma}{\restr}$.  Since $f$ fills an appropriate
argument position, $\IT{scope-arg}=f_\sigma$.  As for $\IT{scope}$,
the scope of a determiner is not explicitly given, so we can only say that
it can be any semantic structure,\footnote{Because of this
indeterminacy as to choice of scope,
\namecite{HalvorsenKaplan:Projections} used {\it inside-out functional
uncertainty}\/ to nondeterministically choose a scope f-structure for
a quantifier.  Their approach requires the scope of a quantifier to be
an f-structure which contains the quantifier f-structure.  In
contrast, our approach places no syntactic constraints whatever on the
choice of quantifier scope, since the propositional structure of the
formulas involved in the derivation will preclude all but the
appropriate choices.}  subject to the constraint that the meaning
associated to the semantic structure have proposition type $t$, that
is, the semantic contribution should quantify universally over
possible scopes.  Therefore, the contribution of a determiner
is\footnote{There is an alternative formulation of quantifier meaning
that doesn't use nested implications: $$\begin{array}{r@{\,}r@{\,}l}
&\exists x. & \attr{\Ups}{\var}\means x \\ & \otimes & \All{ H,R}
\attr{\Ups}{\restr} \means Rx \linimp (\Up_{\sigma} \means x \otimes
\All{S} (H \means Sx)
\linimp H\means Q(z,Rz,Sz))
\end{array}$$

\noindent This formulation just asserts that there is a generic
entity, $x$, which stands for the meaning of the quantified phrase,
and also serves as the argument of the restriction.  The derivations
of the restriction and scope are then expected to consume this
information.  By avoiding nested implications, this formulation may be
easier to work with computationally.

However, the logical structure of this formulation is not as
restrictive as that of (\ref{ex:det-def}), as it can
allow additional derivations where information intended for the
restriction can be used by the scope.  In fact, though, for many
analyses (including all those that we have investigated) such
interactions are already precluded by the quantificational
structure of the formula, and in such cases the formulation above is
equivalent to (\ref{ex:det-def}).  }:
\enumsentence{\label{ex:det-def}
$\begin{array}[t]{r@{\,}l}
\All{ {H}, R, S} & (\All{ x} \attr{\Ups}{\var}
   \means x \linimp \attr{\Ups}{\restr} \means R x)\\
\otimes & (\All{ x} \Ups \means x \linimp {H} \meansub{t} S x)\\
\linimp & {H} \meansub{t} Q(z, Rz, Sz)
\end{array}$}
where ${H}$ ranges over semantic structures.

The $\var$ and $\restr$
components of the semantic projection for a quantified noun phrase in
our analysis play a similar role to the $/\!\!/$ category constructor
in PTQ \cite{Montague:PTQ}, that of distinguishing syntactic
configurations with identical semantic types but different
contributions to the interpretation. The two PTQ syntactic categories
$t/e$ for intransitive verb phrases and $t/\!\!/e$ for common noun
phrases correspond to the single semantic type $e \rightarrow t$;
similarly, the two conjuncts in the antecedent of \pex{ex:det-def}
correspond to the same semantic type, encoded with a linear
implication, but to two different syntactic contexts, one relating the
predication of a noun phrase to its implicit argument and one relating
a clause to an embedded argument.

\subsection{Quantified noun phrase meanings}
\label{sec:qnp-mean}
We first demonstrate how the semantic contribution of a quantified
noun phrase such as {\it every voter} is derived.  The following
annotated phrase structure rule is necessary:

\enumsentence{
\phraserule{NP}{
\rulenode{Det\\ \up = \Down}
\rulenode{N\\ \up = \Down}}}
This rule states that the determiner Det and noun N each contribute to
the f-structure for the NP.  Lexical specifications ensure that the
noun contributes the \pred\ attribute and its value, and the
determiner contributes the \spec\ attribute and its value.  The
f-structure for the noun phrase {\it every voter}\/ is:
\enumsentence{\label{eg:ev-voter}\evnup{
\fd{$h$:\fdand{\feat{\spec}{`every'}
                 \feat{\pred}{`voter'}}}}}
The lexical entries used in this f-structure are:
\enumsentence{
\lexentry{every}{Det}{
\attr{\Up}{\spec} = `every'\\
\hspace*{-3em}$\begin{array}{r@{\,}l}
\All{ {H},R,S} & (\All{ x} \attr{\Ups}{\var}\means x \linimp
\attr{\Ups}{\restr} \means Rx) \\
\otimes & (\All{ x}\Ups \means x \linimp {H} \means Sx) \\
\linimp & {H} \means \IT{every}(z, Rz, Sz)
\end{array}$}}
\enumsentence{
\lexentry{voter}{N}{
\attr{\Up}{\pred} = `voter'\\
$\All{ X}\attr{\Ups}{\var}\means X \linimp \attr{\Ups}{\restr} \means
\IT{voter}(X)$}}
The semantic contributions of common nouns and determiners were
described in the previous section.

Given those entries, the semantic contributions of {\it every} and
{\it voter} in \pex{eg:ev-voter} are
$$\begin{array}{ll}
\BF{every}\colon&
\begin{array}[t]{r@{\,}l}
\All{ {H}, R, S} & (\All{ x} \attr{h_\sigma}{\var} \means x \linimp
\attr{h_\sigma}{\restr} \means Rx) \\
\otimes & (\All{ x}  h_\sigma \means x \linimp {H} \means Sx)\\
\linimp & {H} \means \IT{every}(z, Rz, Sz)
\end{array}\\[2ex]
\BF{voter}\colon&
\All{ X} \attr{h_\sigma}{\var} \means X \linimp
\attr{h_\sigma}{\restr}
\means \IT{voter}(X)
\end{array}$$
\noindent From these two premises, the semantic contribution for {\em
every voter} follows:
$$\begin{array}{lr@{\,}l}
\BF{every-voter}\colon\ &
\All{ {H}, S} & (\All{ x}  h_{\sigma} \means x \linimp {H} \means Sx)\\
&\linimp & {H} \means \IT{every}(z, \IT{voter}(z), Sz)
\end{array}$$
The propositional part of this contribution corresponds to the
standard type for noun phrase meanings, $(e\rightarrow t)\rightarrow
t$. Informally, the whole contribution can be read as follows: if by
giving the arbitrary meaning $x$ of type $e$ to the argument position
filled by the noun phrase we can derive the meaning $S x$ of type $t$
for the semantic structure scope of quantification ${H}$, then
$S$ can be the property that the noun phrase meaning requires as its
scope, yielding the meaning $\IT{every}(z, \IT{voter}(z), Sz)$ for
${H}$. The quantified noun phrase can thus be seen as providing
two contributions to an interpretation: locally, a {\em referential
import} $x$, which must be discharged when the scope of quantification
is established; and globally, a {\em quantificational import} of type
$(e\rightarrow t)\rightarrow t$, which is applied to the meaning of
the scope of quantification to obtain a quantified proposition.

\subsection{Simple example of quantification}

Before we look at quantifier scope ambiguity and interactions between
scope and bound anaphora, we demonstrate the basic operation of our
proposed representation of the semantic contribution of a determiner.
We use the  following sentence with a single quantifier and no
scope ambiguities:
\enumsentence{\label{simple-quant}
Bill convinced every voter.}
To carry out the analysis, we need a lexical entry for
{\em convinced}\/:
\enumsentence{
\lexentry{convinced}{V}{
\attr{\Up}{\pred}= `convince'\\
\attr{\Up}{\tense}= \past\\
\hspace*{-6em}$\All{ X, Y}\attr{\Up}{\subj}_\sigma\means X \otimes
    \attr{\Up}{\obj}_\sigma\means Y \linimp
    \Ups \means \IT{convince}(X, Y)$}}

\noindent The f-structure for (\ref{simple-quant}) is:
\enumsentence{\evnup{
\fd{$f$:\fdand{\feat{\pred}{`convince'}
           \feat{\tense}{\past}
           \feat{\subj}{$g$:\fdand{\feat{\pred}{`Bill'}}}
           \feat{\obj}{$h$:\fdand{\feat{\spec}{`every'}
                                     \feat{\pred}{`voter'}}}}}}}

\noindent The premises for the derivation are the semantic
contributions for {\it Bill}\/ and {\it convinced}\/ together with the
contribution derived above for the quantified noun phrase {\em every
voter}\/:
$$\begin{array}{ll@{\,}l}
\BF{bill}\colon & \makebox[1em][l]{$g_{\sigma} \means Bill$}\\[1ex]
\BF{convinced}\colon & \All{ X, Y} & g_{\sigma} \means X \otimes h_{\sigma}
\means Y\linimp f_{\sigma} \means convince\/(X, Y)\\[1ex]
\BF{every-voter}\colon\ & \All{ {H}, S} &
\/(\All{ x}  h_{\sigma} \means x \linimp {H} \means Sx)\\
 & \hfill \linimp & {H} \means every\/(z, voter\/(z), Sz)
\end{array}$$
Giving the name \BF{bill-convinced} to the formula
\[\begin{array}[t]{ll}
\BF{bill-convinced}\colon& \All{ Y} h_{\sigma} \means Y \linimp  f_{\sigma}
\means convince\/(Bill, Y) \\
\end{array}
\]
we have the derivation:
\[
\hspace*{-1em}\begin{array}{l@{\hspace*{2em}}l}
&\BF{bill} \otimes \BF{convinced} \otimes \BF{every-voter} \quad
\mbox{\/(Premises.)} \\[0.5ex]
\vdash & \BF{bill-convinced} \otimes \BF{every-voter}\\[0.5ex]
\vdash & f_{\sigma} \means \IT{every}(z, \IT{voter}(z), \IT{convince}(Bill, z))
\end{array}
\]
No derivation of a different formula $f_{\sigma} \means_t P$ is
possible.  The formula \BF{bill-convinced} represents the semantics of
the scope of the determiner `every'. The derivable formula \[\All{
Y}h_{\sigma} \meansub{e} Y \linimp h_{\sigma} \meansub{e} Y\] could at
first sight be considered another possible, but erroneous,
scope. However, the type subscripting of the \means\ relation used in
the determiner lexical entry requires the scope to represent a
dependency of a proposition on an individual, while this formula
represents the dependency of an individual on an individual
(itself). Therefore, it does not provide a valid scope for the
quantifier.

\subsection{Quantifier scope ambiguities}
When a sentence contains more than one quantifier, scope ambiguities
are of course possible. In our system, those ambiguities will appear
as alternative successful derivations. We will take as our example the
sentence\footnote{In order to allow for apparent scope ambiguities, we
adopt a scoping analysis of indefinites, as proposed, for example, by
\namecite{Neale:Descriptions}.}
\enumsentence{\label{ae}
Every candidate appointed a manager}
\noindent for which we need the additional lexical entries

\enumsentence{
\lexentry{a}{Det}{
\attr{\Up}{\spec} = `a'\\
$\begin{array}[t]{@{\strut}r@{\,}l@{\strut}}
\All{ {H}, R, S} & (\All{ x}\attr{\Ups}{\var}\means x \linimp
\attr{\Ups}{\restr} \means Rx)  \\
\otimes & (\All{ x} \Ups \means x \linimp {H} \means Sx)  \\
\linimp & {H} \means \IT{a}(z, Rz, Sz)
\end{array}$}}

\enumsentence{
\lexentry{candidate}{N}{
\attr{\Up}{\pred} = `candidate'\\
$\hspace*{-2em}\All{ X} \attr{\Ups}{\var} \means X \linimp \attr{\Ups}{\restr}
\means\IT{candidate}\/(X)$}}

\enumsentence{
\lexentry{manager}{N}{
\attr{\Up}{\pred} = `manager'\\
\hspace*{-2em}$\All{ X}\attr{\Ups}{\var}\means X \linimp \attr{\Ups}{\restr}
\means\IT{manager}\/(X)$}}
The f-structure for sentence (\ref{ae}) is
\enumsentence{\evnup{
\fd{$f$:\fdand{\feat{\pred}{`appoint'}
           \feat{\tense}{\past}
           \feat{\subj}{$g$:\fdand{\feat{\spec}{`every'}
                                     \feat{\pred}{`candidate'}}}
           \feat{\obj}{$h$:\fdand{\feat{\spec}{`a'}
                                     \feat{\pred}{`manager'}}}}}}}

\noindent We can derive semantic contributions
for {\it every candidate}\/ and {\it a manager}\/ in the way shown in
Section \ref{sec:qnp-mean}.  Further derivations proceed from those
contributions together with the contribution of {\em appointed}\/:
$$\begin{array}{lr@{\,}l}
\BF{every-candidate}\colon&
\All{ {H}, S} &(\All{ x}  g_{\sigma}\means x \linimp {H} \means Sx)\\
& \linimp & {H} \means \IT{every}(w, \IT{candidate}(w), Sw)\\[1ex]
\BF{a-manager}\colon& \All{ {H}, S} & (\All{ x}  h_{\sigma} \means x\linimp {H}
\means Sx)\\
& \linimp & {H} \means \IT{a}(z, \IT{manager}(z), Sz)\\[1ex]
\BF{appointed}\colon& \All{ X, Y} & g_{\sigma} \means X\otimes
h_{\sigma} \means Y \linimp f_{\sigma} \means \IT{appoint}(X, Y)
\end{array}
$$

\noindent As of yet, we have not made any commitment about the scopes
of the quantifiers; the $\forall S$'s have not been instantiated.
Scope ambiguities are manifested in two different ways in our system:
through the choice of different semantic structures ${H}$,
corresponding to different syntactic choices for where to scope the
quantifier, or through different relative orders of quantifiers that
scope at the same point.  For this example, the second case is
relevant, and we must now make a choice to proceed. The two possible
choices correspond to two equivalent rewritings of {\bf appointed}:
\[
\begin{array}{l}
\All{ X} g_{\sigma} \means X \linimp (\All{ Y}h_{\sigma} \means Y
\linimp f_{\sigma} \means \IT{appoint}(X, Y)) \\
\All{ Y}h_{\sigma} \means Y\linimp (\All{ X} g_{\sigma} \means X \linimp
f_{\sigma} \means \IT{appoint}(X, Y))
\end{array}
\]
\noindent These two equivalent forms correspond to the two possible
ways of ``currying'' a two-argument function  $f: \alpha\times
\beta\rightarrow \gamma$ as one-argument functions: $$\lambda u.\lambda
v.f(u,v): \alpha \rightarrow (\beta \rightarrow \gamma)$$
$$\lambda v.\lambda
u.f(u,v): \beta \rightarrow (\alpha \rightarrow \gamma)$$
We select {\em a manager} to take
narrower scope by using universal instantiation and
transitivity of implication to combine the
first form with {\bf a-manager} to yield
$$\begin{array}{@{\strut}lr@{\,}l@{\strut}}
\BF{appointed-a-manager}\colon&
 \All{ X} & g_{\sigma}\means X \\
& \linimp & f_{\sigma} \meansub{t} a(z,\IT{manager}\/(z), \IT{appoint}\/(X, z))
\end{array}$$
We have thus the following derivation
\[
\begin{array}{l@{\hspace*{2em}}l}
& \BF{every-candidate} \otimes \BF{appointed} \otimes \BF{a-manager}\\[0.5ex]
\vdash &  \BF{every-candidate} \otimes \BF{appointed-a-manager}\\[0.5ex]
\vdash & f_{\sigma} \meansub{t}\IT{every}\/(w,\IT{candidate}\/(w),
a(z,\IT{manager}\/(z),\IT{appoint}(w, z)))
\end{array}
\]
of the $\forall\exists$ reading of \pex{ae}.

Alternatively, we could have chosen {\em every candidate} to take
narrow scope, by combining the second equivalent form of {\bf appointed}
with {\bf every-candidate} to produce:
\[\begin{array}{ll}
\lefteqn{\BF{every-candidate-appointed}\colon} \\
& \All{ Y} h_{\sigma}\means Y  \linimp f_{\sigma} \meansub{t} \IT{every}\/(w,
\IT{candidate}\/(w), \IT{appoint}(w, Y))
\end{array}
\]

\noindent This gives the derivation
$$
\begin{array}{l@{\hspace*{2em}}l}
& \BF{every-candidate} \otimes \BF{appointed} \otimes \BF{a-manager}\\[0.5ex]
\vdash &  \BF{every-candidate-appointed} \otimes \BF{a-manager}\\[0.5ex]
\vdash & f_{\sigma}
\meansub{t}\IT{a}\/(z,\IT{manager}\/(z),\IT{every}\/(w,\IT{candidate}\/(w),\IT{appoint}(w, z)))
\end{array}
$$ for the $\exists\forall$ reading. These are the only two possible
outcomes of the derivation of a meaning for \pex{ae}, as required. We
have used our implementation to verify that no other outcomes are
possible, since manual verification would be rather laborious.

\subsection{Constraints on quantifier scoping}
\label{sec:constraints}

Sentence (\ref{ex:admirer}) contains two quantifiers and therefore might
be expected to show a two-way ambiguity analogous to the one described
in the previous section:

\enumsentence{\label{ex:admirer}
Every candidate appointed an admirer of his.}

\noindent However, no such ambiguity is found if the pronoun {\it his}
is taken to corefer with the subject {\it every candidate}. In this
case, only one reading is available, in which {\it an admirer of his}
takes narrow scope.  Intuitively, this noun phrase may not take wider
scope than the quantifier {\em every candidate}, on which its
restriction depends.

As we will soon see, the lack of a wide scope {\em a} reading follows
automatically from our formulation of the semantic contributions of
quantifiers without further stipulation. In Pereira's earlier work on
deductive interpretation \cite{Pereira:SemComp,Pereira:HOD}, the same
result was achieved through constraints on the relative scopes of
glue-level universal quantifiers representing the dependencies between
meanings of clauses and the meanings of their arguments. Here,
although universal quantifiers are used to support the extraction of
properties representing the meanings of the restriction and scope (the
variables $R$ and $S$ in the determiner lexical entries), the blocking
of the unwanted reading follows from the propositional structure of
the glue formulas, specifically the nested linear implications. This
is more satisfactory, since it does not reduce the problem of proper
quantifier scoping in the object language to the same problem in the
metalanguage.

The lexical entry for {\it admirer} is:
\enumsentence{
\lexentry{admirer}{N}{
\attr{\Up}{\pred} = `admirer'\\
$\begin{array}[t]{r@{\,}l}
\All{ X, Y} & \attr{\Ups}{\var}\means X
\otimes \attr{\Up}{\mbox{\obl\downlett{\rm OF}}}_\sigma \means Y \\
\linimp & \attr{\Ups}{\restr} \means\IT{admirer}(X, Y)
\end{array}$}}

\noindent Here, {\it admirer} is a relational noun
taking as its oblique argument a phrase with prepositional marker {\it
of}, as indicated in the f-structure by the attribute
\obl\downlett{OF}.  The semantic contribution for a relational noun
has, as expected, the same propositional form as the binary relation
type $e\times e\rightarrow t$: one argument is the admirer, and the
other argument is the admiree.

We assume that the semantic projection for the antecedent of the
pronoun {\it his} has been determined by some separate mechanism and
recorded as the $\ant$ attribute of the pronoun's semantic
projection.\footnote{The determination of appropriate values for
$\ant$ requires a more more detailed analysis of other linguistic
constraints on anaphora resolution, which would need further
projections to give information about, for example, discourse
relations and salience.
\namecite{Dalrymple:SyntaxAnaph} discusses in detail LFG analyses of
anaphoric binding.} The semantic contribution of the pronoun is, then,
a formula that consumes the meaning of its antecedent and then
reintroduces that meaning, simultaneously assigning it to its own
semantic projection:
\enumsentence{
\lexentry{his}{N}{
\attr{\Up}{\pred} = `pro'\\ $\All{ X} \attr{\Ups}{\ant} \means X
\linimp  \attr{\Ups}{\ant} \means X
\otimes \Ups \means X$}}
\noindent In other words, the semantic contribution of a pronoun
copies the meaning $X$ of its antecedent as the meaning of the pronoun
itself.  Since the left-hand side of the linear implication
``consumes'' the antecedent meaning, it must be reinstated in the
consequent of the implication.

\noindent The f-structure for example (\ref{ex:admirer}) is, then:

\enumsentence{\evnup{
\fd{$f$:\fdand{\feat{\pred}{`appointed'}
           \feat{\tense}{\past}
           \feat{\subj}{$g$:\fdand{\feat{\spec}{`every'}
                                     \feat{\pred}{`candidate'}}}
           \feat{\obj}{$h$:\fdand{\feat{\spec}{`a'}
               \feat{\pred}{`admirer'}
               \feat{\obl\downlett{OF}}{$i$:\fdand{\feat{\pred}{`pro'}}}}}}}}}
\noindent with $\attr{i_\sigma}{\ant}=g_\sigma$.

We will begin by illustrating the derivation of the meaning of {\it an
admirer of his}\/, starting from the following premises:
$$\begin{array}[t]{lr@{\,}l}
\BF{a}\colon& \All{ {H}, R, S} & (\All{ x} \attr{h_\sigma}{\var}\means x
  \linimp \attr{h_\sigma}{\restr} \means Rx) \\
& \otimes & (\All{ x}  h_{\sigma} \means x \linimp {H} \means Sx)\\
& \linimp & {H} \means \IT{a}(z, Rz, Sz) \\[1ex]
\BF{admirer}\colon& \All{ Z, X} & \attr{h_\sigma}{\var} \means Z  \otimes
i_{\sigma} \means X\\
& \linimp & \attr{h_\sigma}{\restr} \means \IT{admirer}(Z, X) \\[1ex]
\BF{his}\colon& \All{ X}& g_{\sigma} \means X \linimp \/g_{\sigma} \means X
\otimes
i_{\sigma} \means X
\end{array}$$
First, we rewrite {\bf admirer} into the equivalent form
$$
\All{X} i_{\sigma} \means X  \linimp (\All{Z}\attr{h_\sigma}{\var} \means Z
\linimp \attr{h_\sigma}{\restr} \means \IT{admirer}(Z, X))
$$
\noindent We can use this formula to rewrite the
the second conjunct in the consequent of {\bf his},
yielding
$$\begin{array}[t]{ll}
\lefteqn{\BF{admirer-of-his}\colon} \\
& \begin{array}[t]{ll}\lefteqn{\All{ X} g_{\sigma} \means X \linimp} \\
& g_{\sigma} \means X \otimes \\
& (\All{ Z} \attr{h_\sigma}{\var} \means Z \linimp
\attr{h_\sigma}{\restr} \means \IT{admirer}(Z, X))
\end{array}
\end{array}$$
In turn, the second conjunct in the consequent of {\bf admirer-of-his}
matches the first conjunct in the antecedent of {\bf a} given
appropriate variable substitutions, allowing us to derive
$$\begin{array}{ll}
\lefteqn{\BF{an-admirer-of-his}\colon} \\
&\begin{array}[t]{@{\strut}l@{\strut}l@{\strut}l@{\strut}}
\All{X} & g_{\sigma} \means X\linimp \\
 & g_{\sigma} \means X \otimes
 (\All{ {H}, S} & (\All{ x} h_{\sigma} \means x \linimp
 {H} \means Sx) \linimp \\
 & & {H} \means \IT{a}(z, \IT{admirer}(z, X), Sz))
\end{array}
\end{array}$$
\noindent At this point the other formulas available are:
$$\begin{array}[t]{l}
   \BF{every-candidate}\colon\\
   \quad\quad   \All{ {H}, S}  (\All{ x}  g_{\sigma} \means x \linimp {H}
\means Sx)\\
   \quad\quad \linimp  {H} \means \IT{every}(z, \IT{candidate}(z), Sz)\\[1ex]
   \BF{appointed}\colon \\
   \quad\quad \All{ Z,Y}  g_{\sigma} \means Z  \otimes h_{\sigma}
   \means Y \linimp f_{\sigma} \means \IT{appoint}(Z,Y)
\end{array}$$
We have thus the meanings of the two quantified noun phrases.  The
antecedent implication of \BF{every-candidate} has an atomic
conclusion and hence cannot be satisfied by
\BF{an-admirer-of-his}, which has a conjunctive conclusion.
Therefore, the only possible move is to combine \BF{appointed}
and \BF{an-admirer-of-his}. We do this by first
putting {\bf appointed} in the equivalent form
\[
\All{Z}g_{\sigma} \means Z  \linimp (\All{Y} h_{\sigma} \means Y \linimp
f_{\sigma} \means \IT{appoint}(Z,Y))
\]
\noindent After universal instantiation of $Z$ with $X$, this
can be used to rewrite the first conjunct in the consequent of {\bf
an-admirer-of-his} to derive
\[
\begin{array}[t]{@{\strut}l@{\strut}l@{\strut}}
\All{X} & g_{\sigma} \means X\linimp \\
 & (\All{Y} h_{\sigma} \means Y \linimp f_{\sigma} \means
\IT{appoint}(X,Y))\otimes \\
& (\All{{H}, S} (\All{ x} h_{\sigma} \means x \linimp
 {H} \means Sx) \linimp
 {H} \means \IT{a}(z, \IT{admirer}(z, X), Sz))
\end{array}
\]
Universal instantiation of ${H}$ and $S$ together with modus
ponens with the two conjuncts in the consequent as premises yield
\[ \All{X} g_{\sigma} \means X \linimp f_{\sigma}\meansub{t} a(z,
\IT{admirer}(z,X), \IT{appoint}(X,z))
\]
Finally, this formula can be combined with \BF{every-candidate}
to give the meaning of the whole sentence:
\[
f_{\sigma} \meansub{t} \IT{every}(w,\IT{candidate}(w), a(z,
\IT{admirer}(z,w),\IT{appoint}(w,z)))
\]
In fact, this is the only derivable conclusion, showing that our
analysis blocks those putative scopings in which variables occur
outside the scope of their binders.

\section{Conclusion}

Our approach exploits the f-structure of LFG for syntactic information
needed to guide semantic composition, and also exploits the
resource-sensitive properties of linear logic to express the semantic
composition requirements of natural language.  The use of linear logic
as the glue language in a deductive semantic framework allows a natural
treatment of quantification which automatically gives the right
results for quantifier scope ambiguities and interactions with bound
anaphora.

The analyses discussed here show that our linear-logic encoding of
semantic compositionality captures the interpretation constraints
between quantified noun phrases, their scopes and bound
anaphora. The same basic facts are also accounted for in
other recent treatments of compositionality, in particular categorial
analyses with discontinuous constituency connectives
\cite{Moortgat:discontinuous}.  However, we show elsewhere
\cite{Dalrymple+Lamping+Pereira+Saraswat:intensional} that our
approach has advantages over those accounts, in that certain available
readings of sentences with intensional verbs and quantified noun
phrases that current categorial analyses cannot derive are readily
produced in our analysis.

Recently, \namecite{Oehrle:string} independently proposed a
multidimensional categorial system with types indexed so as to keep
track of the syntax-semantic connections that we represent with
$\means$. Using proof net techniques due to
\namecite{Moortgat:labelled} and \namecite{Roorda:resource},
he maps categorial formulas to first-order clauses similar to our
semantic contributions, except that the formulas arising from
determiners lack the embedded implication. Oehrle's system models
quantifier scope ambiguities in a way similar to ours, but it is not
clear that it can account correctly for the interactions with
anaphora, given the lack of implication embedding in the clausal
representation used.  A full comparison of the two systems is left for
future work.

\section*{Acknowledgments}

We thank Johan van Benthem, Bob Carpenter, Jan van Eijck, Angie
Hinrichs, David Israel, Ron Kaplan, John Maxwell, Michael Moortgat,
John Nerbonne, Stanley Peters, Henriette de Swart and an anonymous
reviewer for discussions and comments.  They are not responsible
for any remaining errors, and we doubt that they will endorse all our
analyses and conclusions, but we are sure that the end result is much
improved for their help.

\appendix
\section{Syntax of the Meaning and Glue Languages}
\label{syn-app}
The meaning language is based on Montague's
intensional higher-order logic. In fact, in the present paper we just
use an extensional fragment with the following syntax:
$$\begin{array}{lrcll}
\mbox{(M-terms)} & M & {::=}  & c & \mbox{(Constants)}\\
& & \alt & x & \mbox{(Lambda-variables)} \\
& & \alt & \lambda x\, M & \mbox{(Abstraction)}\\
& & \alt & M\, M & \mbox{(Application)}\\
& & \alt & X & \mbox{(Glue-language variables)}
\end{array}$$
Terms are typed in the usual way;
logical connectives such as {\em every} and {\em a} are
represented by constants of appropriate type.
For readability, we will often ``uncurry'' $M N_1 \cdots N_m$ as
$M(N_1, \ldots, N_m)$. Note that we allow variables in the glue
language to range over meaning terms.

The glue language refers to three kinds of terms: meaning terms,
f-structures, and semantic or $\sigma$-structures. f- and
$\sigma$-structures are feature structures in correspondence (through
projections) with constituent structure. Conceptually, feature
structures are just functions which, when applied to attributes (a set
of constants), return constants or other feature structures.  In the
following we let $A$ range over some pre-specified set of attributes.
$$\begin{array}{lrcll}
\mbox{(F-terms)} & F & {::=}  & \uparrow & \mbox{(Indexical
reference)}\\
& & \alt & f\; \alt\; g \;\alt\; h\; \alt \cdots & \mbox{(F-structure
constants)} \\
& & \alt & (F A) & \mbox{(Attribute selection)}\\
\quad\\
\mbox{($\sigma$-terms)} & S & {::=}  & F_{\sigma} & \mbox{(Semantic
projection)}\\
& & \alt & (S A) & \mbox{(Attribute selection)}\\
& & \alt & H & \mbox{(Glue-language variable)}
\end{array}$$

Glue-language formulas are built up using linear connectives from
atomic formulas of the form $S \means_{\tau} M$, whose intended
interpretation is that the meaning associated with $\sigma$-structure
$S$ is denoted by term $M$ of type $\tau$. We omit the type subscript
$\tau$ when it can be determined from context.
$$\begin{array}{lrcll}
\mbox{(Glue formulas)} & G & {::=}  &S \means_{\tau} M & \mbox{(Basic
assertion)}\\
& & \alt & G \otimes G & \mbox{(Linear conjunction)}\\
& & \alt & G \linimp G & \mbox{(Linear implication)}\\
& & \alt & \forall X.\,G  & \mbox{(Quantification over M-terms)}\\
& & \alt & \forall H.\,G  & \mbox{(Quantification over $\sigma$-terms)}\\
\end{array}$$

\bibliographystyle{fullname}

\end{document}